\documentclass{article}
\usepackage{graphicx}
\usepackage{tabularx}
\usepackage{multirow}
\usepackage{subfigure}
\usepackage{amsmath}
\title{Auxiliary Learning as a step towards\\ 
Artificial General Intelligence}
\author{Christeen T Jose\\
Vellore Institute of Technology, Vellore, India\\
christeenjosethoomkuzhy@gmail.com}

\renewcommand{\today}{\ifcase \month \or January\or February\or March\or %
April\or May \or June\or July\or August\or September\or October\or November\or %
December\fi, \number \year}
\begin{document}
\maketitle
\begin{abstract}
Auxiliary Learning is a machine learning approach in which the model acknowledges the existence of objects that do not come under any of its learned categories.The name “Auxiliary learning” was chosen due to the introduction of an auxiliary class. The paper focuses on increasing the generality of existing narrow purpose neural networks and also highlights the need to handle unknown objects. The Cat \& Dog binary classifier is taken as an example throughout the paper. 
\end{abstract}
\providecommand{\keywords}[1]{\textbf{\textit{Index terms---}} #1}
\section{Introduction}
Think of a Cat \& Dog Classifier that will not mistake an image of a “wolf” for an image of a “dog” or an image of a “lion” for an image of a “house cat”. The classifier is capable of handling images of such objects which do not belong to either of the pre-defined classes. Here the pre-defined classes being “Cat” and “Dog”. The images handled do not have to be similar to those of pre-defined classes, and in fact they can be images of birds such as a parrot or even of pure backgrounds without any instances of an animal such as a forest or a beach or a city, but our classifier will still be able to handle these images accordingly, even though they do not belong to any of the pre-defined classes. Auxiliary Learning aims to provide classifiers with this capability to distinguish unfamiliar real world objects, similar to how humans are capable of identifying new objects from their surroundings and are able to realise that these objects do not belong to any of the known/learnt classes of objects. Thus our classifier/agent can accept the fact that there are other classes/types of objects in the real world, as opposed to classifying every input image as one of the pre-defined classes.

Through this paper, we aim to introduce a novel approach to building general purpose neural networks. We do not claim that our approach presented in the paper is a complete or perfect solution to building general purpose neural networks but we believe it could be a step in the right direction. Our solution mainly addresses two issues that pose difficulty for the development of a general purpose neural network.  The first issue addressed is of increasing the generality of existing narrow purpose neural networks. The second issue addressed is a possible method for combining multiple narrow purpose neural networks to build a bigger neural network that incorporates the functionalities of all sub neural networks.
\section{Background}
Alan Turing in his 1950 paper mentions that intelligent machines could be taught in the same way a child is normally taught, by pointing out and naming things. He also mentioned that focusing on abstract activities like playing chess was an another approach. \cite{turing2009computing} Artificial Intelligence has evolved greatly over the years to tackle specific problems. IBM's Deep Blue,an expert system, beat world champion Garry Kasparov at chess in 1997. IBM Watson, a question-answering system, won against two of Jeopardy's brightest minds in 2011. \cite{campbell2002deep} 2016 marked a giant leap in the history of Artificial Intelligence when DeepMind's AlphaGo defeated Go world champion Lee Sedol. \cite{silver2016mastering} DeepMind came back with another revolutionary breakthrough in 2020, its AlphaFold had successfully solved the 50-year old research problem of protein folding in biology. \cite{jumper2021highly}

Specialised neural network architectures have emerged in various domains of Artificial Intelligence such as image classification, recommendation systems, image segmentation, reinforcement learning, natural language processing, general adversarial networks, object detection, speech recognition, speech synthesis, etc. Each of these neural networks can perform extremely well in their own domain, but none of them have addressed the issue of building a general purpose neural network. Being able to combine these specialised neural network architectures can be a great advancement in the search for Artificial General Intelligence, as we are able to build from existing models which have given significant results in the past, in their own domains.
\section{Theory}
The problem definition can be better understood through an example. Consider the situation, of a new intelligent agent coming to existence. The agent uses a neural network. The agent thus has the potential to learn, but right now it does not know anything. Now, we are providing the agent with two different sets of images. The first consists of images of dogs playing on the beach while the second consists of cats playing in the garden. We use these two sets of images to train the new agent to classify images of dogs and cats accordingly. Once training has completed, the agent should be capable of classifying new images as belonging to either “dog” class or “cat” class. 

But what if the image provided to the agent for classification does not contain any instances of a dog or a cat? In such cases, the agent will still classify the image as belonging to one of the above mentioned classes. This is because our agent is only aware of two classes of input images. But the world is much larger and cannot fit in the narrow perspective of our new agent. There’s always an infinite amount of new knowledge to be inferred. This infinitely large knowledge can neither be quantified nor be provided in its entirety to our agent for learning. The knowledge can only be learned by our agent with time. Thus we can infer the need for the agent to always retain the potential to learn new classes.

Let’s consider another test case scenario. This time the input image is of a beach. Like our previous image, this image does not contain any instances of a dog or a cat. How will our agent classify this image? There is a higher chance for the agent to classify the given image as belonging to the class “dog”. This is because of its higher similarity to the images that were in the training set for class “dog”, due to the presence of the common beach background. The important point to note here is that our agent is incapable of distinguishing instances of a dog or a cat from its background in the training images. This is clearly visible in this situation due to our choice of dataset. Increasing the variety of backgrounds in every set of images used for training can be a solution to avoiding any biasing of a particular background to a particular class. But the agent will still try to classify the image as belonging to one of only two classes that it is aware of. Thus we can infer the need for a mechanism by which the agent can handle/classify images that do not belong to any of the known classes. An another point to note from this example, is that our agent is incapable of distinguishing the actual objects from their background in the images provided for training, thus coming to the conclusion that the patterns associated with the background are also a property of the object. In this case, the actual object being “dog” and the background being “beach”. Thus we can infer the need for a mechanism by which the agent could distinguish the object from the background in the image.

Now let’s consider a third scenario for prediction/classification. This time the image is of a parrot sitting on a tree. In this scenario we can see how our agent should include a new class namely “parrot” for classifying images, instead of classifying the image as belonging to either “dog” or “cat” class. We can extend this to our previous scenario as well, where the agent should have included the new class “beach”. But there are an infinitely large number of classes that can exist in a real world scenario, so it is not possible to pre-define all of them.

Although we used the Cat \& Dog Classification problem as an example, the solution we are putting forward is not restricted to binary classifiers and can be applied to larger multi-class classifiers as well. We would also like to point out that the solution is not restricted to image data.
\section{Auxiliary Learning}
Auxiliary Learning is a machine learning approach in which the model acknowledge’s the existence of objects/data-points that do not come under any of its learned categories/classes. The name “Auxiliary learning” was chosen due to the introduction of an auxiliary class.

Our solution is to include a new class namely “Others” called the Auxiliary Class, in addition to the pre-defined classes in the neural network for classification. The Auxiliary Class will contain data (images) belonging to a wide variety of classes from real life that do not include any instances of the classes that we are interested in. For example, in case of the Cat \& Dog classifier the Auxiliary Class can hold images that do not belong to both the “Dog” class and the “Cat” class, such as images of ships, parrots, forests, mobile phones, volcanoes, dolphins, radios, cars etc. Thus we can have a neural network with increased generality, as it can accept any input image and produce an output that is logically correct.

But, by the very nature of the Auxiliary Class, our classifier will now suffer from Class Imbalance problem. Class imbalance problem is a direct consequence of the Auxiliary Class having a greater number of data points/ images when compared to the number of images in any of the pre-defined classes. The imbalanced distribution of data will affect the classifiers ability to recognise images belonging to the pre-defined classes. A solution would be to use Weighted Categorical Cross-Entropy (equation \ref{CCE}).

\begin{equation}
\label{CCE}
-\left(w_{p}y log{\left(p\right)}+w_n\left(1-y\right)log{\left(1-p\right)}\right)
\end{equation}

Note that we also need to add a small value; $10^{-7}$, to the predicted values before taking their logs. This is simply to avoid a numerical error that would otherwise occur if the predicted value happens to be zero.

A possible solution to building a general purpose neural network would be to find a means by which we can combine multiple narrow purpose neural networks. Reusing trained models will help to save training time, as larger neural network architectures can take a long time to complete training (depending on the computational resources available). Wouldn't it be great if a single neural network architecture could take an input image and identify cancer or diabetes or the species of an animal or the type of an industrial product and so much more classes? 

A traditional approach to utilise multiple neural networks each from a different domain would be to find a logical hierarchy between the neural networks and to manually program the flow of control between the neural networks as defined by the hierarchy. In the hierarchy each pre-defined class of each neural network can act as the point of succession to the neural network in the next level of the hierarchy. For example the “cat” class of a Cat \& Dog classifier can act as the point of succession for the Cat Breed classifier (a classifier that takes the image of a cat and gives the species of cat as the output). Similarly the “dog” class of a Cat \& Dog classifier can act as the point of succession for the Dog Breed classifier (a classifier that takes the image of a dog and gives the species of dog as the output).

But the traditional approach described above will fail when provided with neural networks that come from unrelated domains and where a hierarchical relationship cannot be defined. The use of Auxiliary classes can be a solution to building a system of unrelated neural networks. Auxiliary classes can help to combine systems of neural networks where each system has its own hierarchy. The Auxiliary classes from two different neural networks can come together to form a point of fusion between these neural networks, thus also enabling the  combination of the systems of neural networks they each belong to if any. In a system with multiple neural networks, the Auxiliary class of each neural network presents a possibility towards forming a point of fusion with other systems. The point of fusion can be described as a common point of generality between the involved neural network systems, thus enabling us to combine the functionalities of each system and its components to build a bigger general purpose neural network.

Although the paper talks about neural networks, the concepts that have been presented here are applicable to all classification algorithms.

\section{Implementation}
Since I used the example of the Cat \& Dog Classifier throughout the paper, I decided to conduct my study on the same classifier through the introduction of an Auxiliary class.
\subsection{Datasets}
In order to build the data points for the Auxiliary Class I decided to use the Imagenet dataset. The most highly-used subset of ImageNet is the ImageNet Large Scale Visual Recognition Challenge (ILSVRC) 2012-2017 image classification and localization dataset. This dataset spans 1000 object classes and contains 1,281,167 training images, 50,000 validation images and 100,000 test images. This subset is available on Kaggle and was used for this paper. \cite{russakovsky2015imagenet}

To build the data points for “Cat” and “Dog” classes I used the dataset provided as part of the Dogs vs. Cats competition on Kaggle. The dataset contains 25,000 images of dogs and cats for training. \cite{elson2007asirra}

The ILSVRC dataset had images from 120 species of dogs and few species of cats. These images had to be removed from the dataset before we could build the auxiliary class as the ILSVRC dataset was to be used solely for building the Auxiliary Class (“Others” class). The names of the 120 breeds of dogs in the ILSVRC dataset were provided in the Stanford Dogs Dataset. \cite{khosla2011novel} Before I could remove the images belonging to these categories I had to change the format of the breed names from dash separated to camel case and map the transformed text with the help of LOC\_synset\_mapping.txt file provided as part of the ILSVRC dataset. The LOC\_synset\_mapping.txt file defines the mapping between each of the 1000 synset ids and their corresponding descriptions.

875 classes from ILSVRC dataset was used to build the Auxiliary Class. The number of classes, 875, is obtaining after removal of 120 breeds of dogs and 5 breeds of cats from the 1000 classes of ILSVRC dataset. The 5 breeds of cats present in the ILSVRC dataset were identified manually and they are 'tabby', 'tiger cat', 'Persian cat', 'Siamese cat' and 'Egyptian cat'.
\subsection{Experimental setup}
I used only a subset of the ILSVRC dataset (1 lakh images out of a total 12 lakh images). This was done in order to prevent my laptop from crashing due to Out Of Memory error as I was constrained to a RAM size of 16GB (The ILSVRC dataset was 155 GB in size).

The dataset was further split into train and test sets with 80\% of the images being used for training. Table \ref{Distribution} shows the distribution of data.

\begin{table}[ht]
\begin{center}
\caption{Distribution of data}
\label{Distribution}
\begin{tabular}{|c|c|c|}
\hline
Class  & No. of images used for training & No. of images used for testing\\\hline
Cat    & 9999  ($\sim$10000)    & 2501 $\sim$2500)\\\hline
Dog    & 10001 ($\sim$10000)    & 2499 ($\sim$2500)\\\hline
Others & 87500 (8.75 x 10000)   & 21875 (8.75 x 2500)\\\hline
\end{tabular}
\end{center}
\end{table}

Since the train set had images in the ratio 1:1:8.75, only 21875 images were chosen from a total of 43750 images from Auxiliary Class test set, in order to maintain the ratio for Weighted Categorical Cross-Entropy. It is important to note that, the model can achieve an accuracy of 81.39\%, just predicting “Others” for every image. 

\begin{figure}
    \centering
    \includegraphics[width=\textwidth]{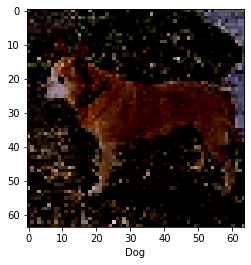}
    \caption{An image of a dog after preprocessing}
    \label{preProcessedDog1}
\end{figure}

The images were further resized to 64 x 64 and pixel values were scaled to be between -1 and 1 and the corresponding labels were one hot encoded. Figure \ref{preProcessedDog1} shows an image of a dog obtained after preprocessing.

It was decided to make use of transfer learning in order to save computational time and yield quicker results. The model chosen for transfer learning was MobileNet. \cite{howard2017mobilenets}

Weights for Weighted Categorical Cross-Entropy:

\begin{equation}
\begin{split}
\text{positive weights}, w_p &= [9.75/10.75, 9.75/10.75, 2/10.75]\\
w_p &= [0.906976744, 0.906976744, 0.186046511]\\
\text{negative weights}, w_n &= 1 - w_p
\end{split}
\end{equation}

\newpage
\section{Findings and analysis}
The experiment was conducted twice,initially with Weighted Categorical \\Cross-Entropy and afterwards with Categorical Cross-Entropy as loss function. Table \ref{TestReultsAuxiliaryLearning} summarises our findings.

\begin{table}[ht]
\begin{center}
\caption{Test Results for Cat \& Dog Classifier with Auxiliary Learning}
\label{TestReultsAuxiliaryLearning}
\begin{tabular}{|c|c|c|}
\hline
Loss function               & Test Accuracy & Test Loss \\\hline
Categorical Cross-Entropy   & 0.95880       & 0.76609   \\\hline
Weighted Categorical Cross-Entropy & 0.94783& 0.24978   \\\hline
\end{tabular}
\end{center}
\end{table}

Thus we have successfully implemented Auxiliary Learning, and managed to achieve significant Test accuracies of 96\% and 95\% with categorical cross-entropy loss and weighted categorical cross-entropy loss, respectively. Although we obtained similar performances with both loss functions, we believe that the use of Weighted Categorical Cross Entropy becomes more profound when the ratio of the total number of images in auxiliary class to the total number of images in any of the predefined classes is much larger.

The experiment was repeated for the normal Cat \& Dog Binary Classifier. The model was adjusted to have only two output classes namely "Cat" \& "Dog", instead of three.  We used two classes for consistency even though binary classification could be implemented with just one class. Categorical cross entropy was used since the data was well balanced amongst the two classes. 20,000 images were used for training and 5000 images were used for testing. Table \ref{TestReultsBinaryCLassifier} summarises our findings.

\begin{table}[ht]
\begin{center}
\caption{Test Results for Cat \& Dog Binary Classifier without Auxiliary Learning}
\label{TestReultsBinaryCLassifier}
\begin{tabular}{|c|c|c|}
\hline
Model                           & Test 			Accuracy & Test 			Loss \\\hline
Cat 			\& Dog Binary Classifier & 0.90939          & 0.78215    \\\hline 
\end{tabular}
\end{center}
\end{table}

\begin{figure}
    \centering
    \subfigure[Confusion matrix for Auxiliary Learning with Weighted Categorical Cross-Entropy]{
        \includegraphics[width=\linewidth]{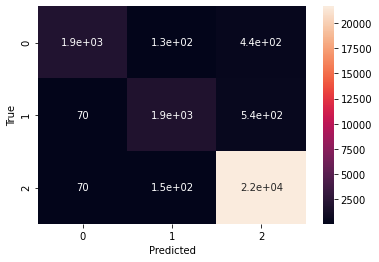}
    }
    \subfigure[Confusion matrix for Cat \& Dog Binary Classifier]{
        \includegraphics[width=\linewidth]{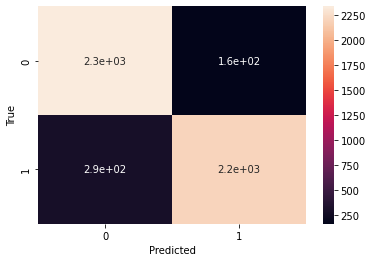}
    }
    \caption{Confusion Matrices}
    \label{CMs}
\end{figure}

Table \ref{PRF1} records precision ,recall and F1-score observed for all classes during all 3 experiments. Confusion matrices have been provided in Figure \ref{CMs}. The confusion matrix for Auxiliary Learning with Categorical Cross Entropy is not included as it is similar to the one for Auxiliary Learning with Weighted Categorical Cross-Entropy.

\begin{table}[ht]
\begin{center}
\caption{Precision ,Recall and F1-Score}
\label{PRF1}
\begin{tabular}{|c|c|c|c|c|}
\hline
Class                   & Model                                                 & Precision & Recall & F1-Score \\\hline
\multirow{3}{*}{Cat}    & Binary 			Classifier                                  & 0.89      & 0.94   & 0.91     \\\cline{2-5}
                        & Auxiliary Learning                    & 0.90      & 0.85   & 0.88     \\\cline{2-5}
                        & AL with weighted loss & 0.93      & 0.77   & 0.84     \\\hline
\multirow{3}{*}{Dog}    & Binary 			Classifier                                  & 0.93      & 0.88   & 0.91     \\\cline{2-5}
                        & Auxiliary Learning                    & 0.87      & 0.83   & 0.85     \\\cline{2-5}
                        & AL with weighted loss & 0.87      & 0.76   & 0.81     \\\hline
\multirow{3}{*}{Others} & Binary 			Classifier                                  & NA        & NA     & NA       \\\cline{2-5}
                        & Auxiliary Learning                    & 0.97      & 0.99   & 0.98     \\\cline{2-5}
                        & AL with weighted loss & 0.96      & 0.99   & 0.97   \\\hline 
\end{tabular}
\end{center}
\end{table}

\newpage
\section{Discussion}
This is the first time a large amount of unrelated data has been put together under a single class. Future scope of the project (apart from Artificial General Intelligence) includes application in bio-metric authentication systems where it is essential for the system to distinguish between authorised users and imposters.
\section*{Acknowledgment}
The project was initially completed as part of CSE3013 (Artificial Intelligence) course at Vellore Institute of Technology (VIT) in partial fulfilment for the award of the degree of Bachelor of Technology in Computer Science and Engineering with Specialization in Bioinformatics. 

I would also like to express my gratitude to my guide, Prof. Geraldine Bessie Amali, for her appreciation and support.
\bibliographystyle{unsrt}
\bibliography{refs}
\end{document}